\documentclass[runningheads,a4paper]{llncs}

\usepackage{amssymb}
\usepackage{graphicx}
\usepackage{amsmath}
\usepackage{amssymb}
\usepackage{subfigure}
\usepackage{enumerate}
\usepackage{color}
\usepackage{bm}
\usepackage{epsfig}
\usepackage{subfigure}
\usepackage{multirow}
\usepackage{array}
\usepackage{microtype}
\usepackage{amsmath,bm}
\usepackage{epstopdf}
\usepackage{cite}
\usepackage{booktabs}

\usepackage{url}

\makeatletter
\newcommand*{\textoverline}[1]{$\overline{\hbox{#1}}\m@th$}
\makeatother

\newcommand{\ie}{\textit{i}.\textit{e}. }

\usepackage{url}
\urldef{\mailsa}\path|{alfred.hofmann, ursula.barth, ingrid.haas, frank.holzwarth,|
\urldef{\mailsb}\path|anna.kramer, leonie.kunz, christine.reiss, nicole.sator,|
\urldef{\mailsc}\path|erika.siebert-cole, peter.strasser, lncs}@springer.com|    
\newcommand{\keywords}[1]{\par\addvspace\baselineskip
\noindent\keywordname\enspace\ignorespaces#1}

\begin{document}

\mainmatter  

\title{CT Image Enhancement Using Stacked Generative Adversarial Networks and Transfer Learning for Lesion Segmentation Improvement}

\titlerunning{CT Image Enhancement Using Stacked GANs and Trasfer Learning}

%
%
\author{Youbao Tang\inst{1}\thanks{indicates equal contribution}, Jinzheng Cai\inst{1,2}\protect\footnotemark[1], Le Lu\inst{1}, Adam P. Harrison\inst{1}, Ke Yan\inst{1}, \\
	Jing Xiao\inst{3}, Lin Yang\inst{2}, Ronald M. Summers\inst{1} \\
	\institute{National Institutes of Health Clinical Center, Bethesda, MD 20892, USA \\
		\email{youbao.tang@nih.gov}
		\and University of Florida, Gainesville, FL 32611, USA
		\and Ping An Insurance Company of China, Shenzhen, 510852, China
	}
}

\authorrunning{Youbao Tang, \textit{et al}.}
\maketitle

\begin{abstract}
Automated lesion segmentation from computed tomography (CT) is an important and challenging task in medical image analysis. While many advancements have been made, there is room for continued improvements. One hurdle is that CT images can exhibit high noise and low contrast, particularly in lower dosages. To address this, we focus on a preprocessing method for CT images that uses stacked generative adversarial networks (SGAN) approach. The first GAN reduces the noise in the CT image and the second GAN generates a higher resolution image with enhanced boundaries and high contrast. To make up for the absence of high quality CT images, we detail how to synthesize a large number of low- and high-quality natural images and use transfer learning with progressively larger amounts of CT images. We apply both the classic GrabCut method and the modern holistically nested network (HNN) to lesion segmentation, testing whether SGAN can yield improved lesion segmentation. Experimental results on the DeepLesion dataset demonstrate that the SGAN enhancements alone can push GrabCut performance over HNN trained on original images. We also demonstrate that HNN + SGAN performs best compared against four other enhancement methods, including when using only a single GAN. 

\keywords{CT image enhancement, lesion segmentation, stacked generative adversarial networks, transfer learning}
\end{abstract}

\section{Introduction}
\begin{figure}[t!]
	\begin{center}
		\subfigure[]{\includegraphics[height=0.5\linewidth]{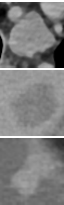}}
		\subfigure[]{\includegraphics[height=0.5\linewidth]{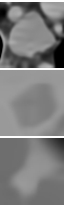}}
		\subfigure[]{\includegraphics[height=0.5\linewidth]{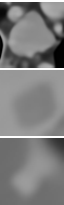}}
		\subfigure[]{\includegraphics[height=0.5\linewidth]{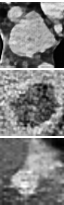}}
		\subfigure[]{\includegraphics[height=0.5\linewidth]{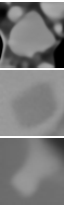}}
		\subfigure[]{\includegraphics[height=0.5\linewidth]{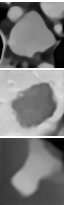}}
	\end{center}
	\caption{Three examples of CT image enhancement results using different methods on original images (a), BM3D (b), DnCNN (c), single GAN (d), our denoising GAN (e), and our SGAN (f).}
	\label{fig:example}
\end{figure}

There are many useful and important applications in medical image analysis, \textit{e.g.}, measurement estimation \cite{tang2018semi}, lung segmentation \cite{jin2018ct}, lesion segmentation \cite{cai2018accurate}, etc. Accurate lesion segmentation from computed tomography (CT) scans plays a crucial role in computer aided diagnosis (CAD) tasks , \textit{e.g.}, quantitative disease progression, tumor growth evaluation after treatment, pathology detection and surgical assistance. Quantitative analysis of tumor extents could provide valuable information for treatment planning. Manual lesion segmentation is highly tedious and time consuming, motivating a number of works on automatic lesion segmentation~\cite{massoptier2008new,christ2016automatic,cai2018accurate}. However, as more and more elaborately designed segmentation methods are proposed, performance improvement may plateau. In particular, CT images are often noisy and suffer from low contrast due to radiation dosage limits, as shown in the first row of Fig. \ref{fig:example}. The collection of datasets more massive than currently available may provide the means to overcome this, but this eventuality is not guaranteed, particularly given the labor involved in manually annotating training images. We take a different tack, and instead leverage the massive amounts of data already residing in hospital picture archiving and communication systems (PACS) to develop a method to enhance CT images in a way that benefits lesion segmentation.

Fig. \ref{fig:example} presents some examples of current efforts at image enhancement. As Fig. \ref{fig:example}(a) demonstrates, classic denoising methods, such as BM3D \cite{dabov2007image}, can preserve image details while introducing very few artifacts. With the recent explosive development of deep convolutional neural networks (CNNs), the field has developed many CNN based denoising methods. These include DnCNN~\cite{zhang2017beyond}, which is able to handle denoising with unknown noise levels. However, most of the CNN based methods, including DnCNN~\cite{zhang2017beyond}, use mean squared error (MSE) loss for model optimization, which can blur high-frequency details, \textit{e.g.} edges. See Fig. \ref{fig:example}(c) for an example. Moreover, denoising methods do not explicitly address resolution and contrast issues.

To overcome these problems, this paper proposes a novel CT image enhancement method by designing a stacked generative adversarial network (SGAN) model. As such, this work builds off of classic GANs~\cite{goodfellow2014generative}, and is partially inspired by work using GANs for super resolution on natural images~\cite{ledig2016photo}. Unlike many natural images, CT images are often noisy and suffer from low contrast. Directly enhancing such images may generate undesirable visual artifacts and edges that are harmful for lesion segmentation accuracy. It is challenging to train a single GAN to directly output  enhanced images with high resolution and visual quality from the original CT images.  See Fig. \ref{fig:example}(d) for the results produced by single GAN. One way to address this is to reduce CT image noise before image enhancement.  Therefore, our proposed SGAN operates in two GAN stages. As shown in Fig.~\ref{fig:example}(e), the first GAN reduces the noise from the original CT image. As depicted in Fig.~\ref{fig:example}(f), the second GAN generates higher resolution images with enhanced boundary and contrast. Based on the enhanced images, the popular segmentation methods of GrabCut and holistically nested networks (HNNs) are used for lesion segmentation. Experimental results on the large scale DeepLesion dataset~\cite{yan2018deep} demonstrate the effectiveness of our SGAN approach.  In particular, we demonstrate that when using SGAN-enhanced with GrabCut, we can produce better results than the much more powerful, yet expensive, HNN applied to the original images, confirming our intuition on the value of attending to image quality.

\begin{figure}[t!]
	\begin{center}
		\includegraphics[width=0.99\linewidth]{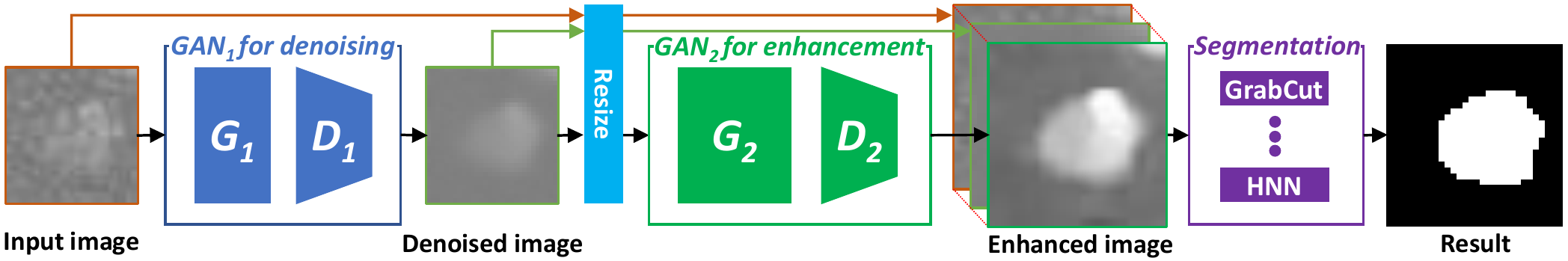}
	\end{center}
	\caption{The pipeline of the proposed method.}
	\label{fig:framework}
\end{figure}

\section{Methods}
Instead of directly performing image enhancement, our SGAN method decomposes enhancement into two sub-tasks, \textit{i.e.}, image denoising followed by enhancement. After SGAN enhancement, either GrabCut or HNN is used for lesion segmentation. Figure~\ref{fig:framework} depicts the overall workflow of the proposed method. The details of each stage are described below.

\subsection{CT Image Enhancement}
In~\cite{ledig2016photo}, generative adversarial networks (GANs)~\cite{goodfellow2014generative} were successfully used for natural-image super resolution,  producing high-quality images with more visual details and edges compared to their low-resolution counterparts. For lesion segmentation, if we can improve visual clarity and contrast, particularly at the borders of lesions, the segmentation performance and accuracy may subsequently be improved. 

Given a CT lesion image (as shown in Fig.~\ref{fig:example}(a)), we first generate a denoised version of the input image by employing our first GAN model (consisting of a generator {\small $G_1$} and a discriminator {\small $D_1$}) that focuses on removing random image noise. The denoised image has the same size as the input image. Although the noise has been reduced in the generated image, as demonstrated in Fig.~\ref{fig:example}(e), lesions have blurry edges and the contrast between lesion and background regions is generally low. As well, a considerable number of lesions are quite small in size ($<10$mm or less than 10 pixels according to their long axis diameters). Human observers typically apply zooming (via commercial clinical PACS workstations) for such lesions. This motivates the use of a second GAN to provide high-resolution enhancement. To solve this issue, our second GAN model, which also contains a generator {\small $G_2$} and a discriminator {\small $D_2$}, is built upon the denoised image from the first GAN to produce an enhanced high resolution version (as illustrated in Fig.~\ref{fig:example}(f)). This enhanced high-resolution image provides both clear lesion boundaries and high contrast. Since the three resulting images, \ie{}, the original, denoised, and enhanced variants, may have complementary information, we concatentate them together into a three-channel image that is fed into the next lesion segmentation stage.

\subsubsection{SGAN Architecture}
We adapt similar architectures as~\cite{ledig2016photo} for the generators and discriminators, where the generator has 16 identical residual blocks and 2 sub-pixel convolutional layers~\cite{shi2016real}, which are used to increase the resolution. Each block contains two convolutional layers with 64 $3 \times 3$ kernels followed by batch-normalization~\cite{ioffe2015batch} and ParametricReLU~\cite{he2015delving} layers. Because it is an easier subtask, a simpler architecture that contains just 9 identical residual blocks is designed for the denoising generator {\small $G_1$}. As well, for a trained model, the method of~\cite{ledig2016photo} can only enlarge the input image by fixed amounts. However, in the DeepLesion dataset lesion sizes vary considerably, meaning they have to be enlarged with correspondingly different zooming factors. Therefore the sub-pixel layers are removed in the high-resolution generator {\small $G_2$}.  Both {\small $G_1$} and {\small $G_2$} are fully convolutional and can take input images of arbitrary size. For the discriminator design, {\small $D_1$} and {\small $D_2$}, we use the same architecture as~\cite{ledig2016photo}, which consists of 8 convolutional layers with $3\times3$ kernels, LeakyReLU activations {\small $(\alpha=0.2)$}, and two densely connected layers followed by a final sigmoid layer. The stride settings and kernel numbers of the 8 convolutional layers are {\small $(1,2,1,2,1,2,1,2)$} and {\small $(64,64,128,128,256,256,512,512)$}, respectively.

\subsubsection{Training Data Synthesization}
Normally super resolution models are trained with pairs of low- and high-resolution images. While this can be obtained easily in natural images (by down-sampling), physical CT images are imaged by medical scanners at roughly fixed in-plane resolutions of $\sim1$ mm per-pixel and CT imaging at ultra-high spatial resolutions does not exist. 
For the sake of SGAN training, we leverage transfer learning using a large-scale synthesized natural image dataset: DIV2K \cite{agustsson_2017_cvpr_workshops} where all images are converted into gray scale and down-sampled to produce training pairs. For the training of the denoising GAN, we randomly crop {\small $32\times32$} sub-images from distinct training images of DIV2K. White Gaussian noise at different intensity variance levels {\small $\sigma_i\in(0,50]$} are added to the cropped images to construct the paired model inputs. For training the image-enhancement GAN, the input images are cropped as {\small $128\times128$} patches and we perform the following steps: 1) down-sample the cropped image with scale {\small $s\in[1,4]$}, 2) implement Gaussian spatial smoothing with {\small $\sigma_s\in(0,3]$}, 3) execute contrast compression with rates of {\small $\kappa\in[1,3]$}, and 4) conduct up-sampling with the scale {\small $s$} to generate images pairs. To fine-tune using CT images, we process $28,000$ training RECIST slices using the currently trained SGAN and select a subset of up to $1,000$ that demonstrate visual improvement. The selected CT images are subsequently added to the training for the next round of SGAN fine-tuning. This iterative process finishes when no more visual improvement can be observed.

\subsubsection{Model Optimization}
A proper loss function needs to be defined for model optimization, which is critical for the performance of our generators. The generators are trained not only to generate high quality images but also to fool the discriminators. Similar to~\cite{ledig2016photo}, given an input image, $\mathbf{x}_{i} \ (i=0,1)$, this work defines a perceptual loss $ L_P^i \ (i=1,2) $ as the weighted sum of a image content loss $ L_C^i $, a feature representation loss $L_{VGG}^i$ and an adversarial loss $ L_A^i $ for $ G_1 $ and $ G_2 $ as
\begin{equation}
L_P^i = L_{DIFF}^i + 10^{-5}L_{VGG}^i + 10^{-3}L_A^i \textrm{,}
\end{equation}
where $i$ denotes the SGAN stage. Here, $ L_{DIFF}^i $ and $L_{VGG}^i$ are computed using the mean square error (MSE) loss function to measure the pixel-wise error and the element-wise error of feature maps between the generated image $ G_i(\mathbf{x}_i) $ and its ground truth image $ \mathbf{y}_i $, respectively. We extract feature maps from five blocks of the VGGNet-16 model \cite{simonyan2014very} pre-trained over ImageNet \cite{deng2009imagenet}. The adversarial loss $ L_A^i $ is defined using the standard GAN formulation for generators:
\begin{equation}
L_A^i(\mathbf{x}_i) =- \log(D_i(G_i(\mathbf{x}_i))) \textrm{.}
\end{equation}

The discriminators, on the other hand, are trained to distinguish between real images and enhanced ones, $\mathbf{y}_i$ and $ G_i(\mathbf{x}_i) $, respectively,which can be accomplished by minimizing the following loss:
\begin{equation}
L_D^i(\mathbf{x}_i,\mathbf{y}_i) = -\log(D_i(\mathbf{y}_i)) - \log(1 - D_i(G_i(\mathbf{x}_i)))
\end{equation}

We use the Adam optimizer~\cite{kingma2014adam} with $ \beta_1=0.5 $ and a learning rate of $ 10^{-4} $ for model optimization. The generator ($ G_1 $ or $ G_2 $) and discriminator ($ D_1 $ or $ D_2 $) are alternatively updated. We train first on the synthesized natural images and then fine-tune using the selected CT images.

\subsection{Lesion Segmentation}
Because they may contain complementary information, the denoised and enhanced outputs from the SGAN and the original lesion CT image are combined into a three-channel image for lesion segmentation. We investigate two popular segmentation approaches: GrabCut~\cite{rother2004grabcut} and HNN~\cite{xie2015holistically}. The quality of GrabCut's initialization will greatly affect the final segmentation result. For this reason, we construct a high quality  \textit{trimap} {\small $T$} using the RECIST diameter marks within the DeepLesion dataset~\cite{yan2018deep}. This produces regions of probable background, probable foreground, background and foreground. Note that unlike the original \textit{trimap} definition~\cite{rother2004grabcut}, we define four region types. With {\small $T$}, we can obtain the lesion segmentation using GrabCut. Since the DeepLesion dataset does not provide the ground truth lesion masks, the GrabCut segmentation results are used as supervision to  train the HNN segmentation model until convergence.

\section{Experimental Results and Analyses}
The DeepLesion dataset \cite{yan2018deep} is composed of $32,735$ PACS CT lesion images annotated with RECIST long and short diameters. These are derived from $10,594$ studies of $4,459$ patients. All lesions have been categorized into the $8$ subtypes of lung, mediastinum, liver, soft-tissue, abdomen, kidney, pelvis, and bone. For quantitative evaluation, we manually segment $1,000$ lesion images as a testing set, randomly selected from 500 patients. The rest serve as a training set. Based on the location of bookmarked diameters, CT region of interests (ROIs) are cropped at two times the extent of the lesion's long diameters, so that sufficient visual context is preserved. Although we do not possess corresponding high quality images in the DeepLesion dataset, we can implicitly evaluate the performance of the proposed SGAN model for CT image enhancement by comparing the segmentation performance with or without enhanced images. Three criteria, \ie Dice similarity coefficient (Dice), precision and recall scores, are used to evaluate the quantitative segmentation accuracy. 

\begin{figure}[t!]
	\begin{center}
		\includegraphics[width=0.99\linewidth]{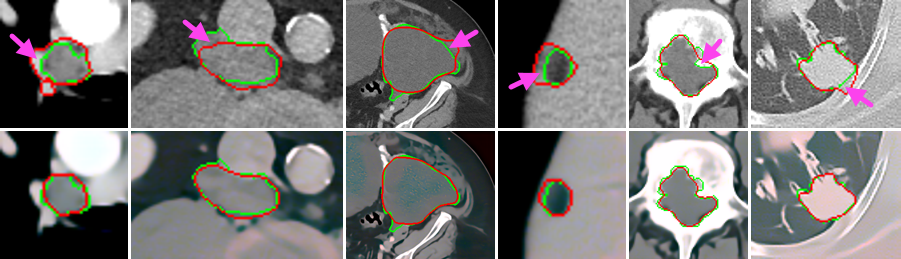}
	\end{center}
	\caption{Visual examples of lesion segmentation produced by HNN from the OG images ($1^{st}$ row) and their combinations with enhanced SGAN images ($2^{nd}$ row). The manual and automatic segmentation boundaries are delineated with green and red curves, respectively. Incorrectly segmented regions when using OG images that are corrected when using SGAN images are highlighted with pink arrows. Best viewed in color.}
	\label{fig:visual-example}
\end{figure}

Figure \ref{fig:visual-example} shows several visual examples of lesion segmentation results using HNN on original images and their combinations with the images produced by SGAN. From Fig. \ref{fig:visual-example}, the segmentation results on the combined images are closer to the manual segmentations than the ones on only original images. This intuitively demonstrates that the enhanced images produced by the SGAN model is helpful for lesion segmentation.

\begin{table}[t!]
	\caption{The performance of lesion segmentation using GrabCut and HNN with different inputs in terms of recall, precision and Dice score, whose mean and standard deviation are reported.}
	\linespread{1.2}
	\scriptsize
	\begin{center}
		\begin{tabular}{|@{}*{1}{m{1.85cm}<{\centering}@{}|@{}}*{6}{m{1.69cm}<{\centering}@{}|@{}}}
			\hline
			\multirow{2}*{\textbf{Input}}  & \multicolumn{3}{c|}{\textbf{GrabCut} } & \multicolumn{3}{c|}{\textbf{HNN}} \\ \cline{2-7}
			&  \textbf{Recall} &  \textbf{Precision}  &  \textbf{Dice}  &  \textbf{Recall} &  \textbf{Precision}  &  \textbf{Dice}  \\\hline
			\textbf{OG}  &\textbf{0.944$\pm$0.096} &0.885$\pm$0.107 &0.908$\pm$0.088 &0.933$\pm$0.095 &0.893$\pm$0.111 &0.906$\pm$0.089 \\ 
			\textbf{OG+BM3D}  &0.943$\pm$0.105 &0.897$\pm$0.105 &0.910$\pm$0.087 &0.903$\pm$0.108 &0.930$\pm$0.095 &0.912$\pm$0.085 \\ 
			\textbf{OG+DnCNN}  &0.944$\pm$0.101 &0.892$\pm$0.108 &0.909$\pm$0.090 &0.901$\pm$0.114 &0.927$\pm$0.098 &0.910$\pm$0.086 \\ 
			\textbf{OG+GAN}  &0.944$\pm$0.107 &0.878$\pm$0.112 &0.906$\pm$0.093 &\textbf{0.937$\pm$0.109} &0.887$\pm$0.108 &0.906$\pm$0.091 \\ 
			\textbf{OG+\bm{$ \mathrm{GAN_1} $}}  &0.942$\pm$0.102 &0.898$\pm$0.106 &0.910$\pm$0.086 &0.905$\pm$0.104 &0.930$\pm$0.093 &0.913$\pm$0.084 \\ 
			\textbf{OG+SGAN}  &0.941$\pm$0.106 &\textbf{0.904$\pm$0.096} &\textbf{0.913$\pm$0.085} &0.911$\pm$0.097 &\textbf{0.940$\pm$0.091} &\textbf{0.920$\pm$0.082} \\ \hline 
		\end{tabular}
	\end{center}
	\label{table:results-input}
\end{table}

\begin{table}[t!]
	\begin{center}
		\caption{Category-wise comparisons of lesion segmentation results using HNN on original images and their combinations with the images produced by SGAN. Mean and standard deviation of Dice score are reported.}
		{
			\scriptsize
			\linespread{1.2}
			
			\begin{tabular}{|@{}*{1}{m{2.3cm}<{\centering}@{}|@{}}*{5}{m{1.9cm}<{\centering}@{}|}}
				\hline
				\textbf{Method} & \textbf{bone} & \textbf{abdomen} & \textbf{mediastinum} & \textbf{liver} & \\ \hline
				\textbf{HNN}     & 0.877$\pm$0.055  & 0.909$\pm$0.092  & 0.892$\pm$0.076  & 0.854$\pm$0.146  & \\
				\textbf{SGAN+HNN}  & \textbf{0.891$\pm$0.061}  & \textbf{0.927$\pm$0.088}  & \textbf{0.909$\pm$0.083}  & \textbf{0.877$\pm$0.142}  & \\ \hline
				\hline
				\textbf{Method} & \textbf{lung} & \textbf{kidney} & \textbf{soft tissue} & \textbf{pelvis} & \textbf{mDice} \\ \hline
				\textbf{HNN}     & 0.912$\pm$0.087  & 0.925$\pm$0.056  & 0.928$\pm$0.063  & 0.911$\pm$0.070  & 0.906$\pm$0.089 \\
				\textbf{SGAN+HNN}  & \textbf{0.924$\pm$0.073}  & \textbf{0.938$\pm$0.045}  & \textbf{0.937$\pm$0.048}  & \textbf{0.919$\pm$0.080}  & \textbf{0.920$\pm$0.082} \\ \hline
			\end{tabular}
		}
	\end{center}
	\label{tab:results-categary}
\end{table}

For quantitative evaluation, we test both GrabCut and HNN applied on a variety of image options. Namely, we compare results when the original (OG) images are used, and also when those processed by BM3D~\cite{dabov2007image}, DnCNN~\cite{zhang2017beyond}), a single GAN model (GAN) for enhancement, and the first denoising GAN model of SGAN ($ \mathrm{GAN_1} $) are used. When enhancement is applied, we concatenate the result with OG images to create a multi-channel input for the segmentation method. From Table \ref{table:results-input}, we can see that 1) when using any of the enhanced images except the one produced by a single GAN, the Dice performance improves, supporting our intuition on the value of enhancing the CT images prior to segmentation. The possible reason of the worse results when using a single GAN is that it may enhance and introduce some artifacts. 2) Using $ \mathrm{GAN_1} $ for image denoising produces better Dice scores than using BM3D and DnCNN, suggesting that the adversarial learning strategy is helpful for image denoising while keeping details important for segmentation. 3) Compared with GrabCut, HNN achieves a greater improvement in Dice scores when using the enhanced images, suggesting HNN is able to exploit the complementary information better than GrabCut. 4) Using SGAN for image enhancement produces the largest gain in Dice scores. This confirms that our approach on using a stacked architecture can provide a more effective enhancement than just a blind application of GANs. 5) Most remarkably, the Dice scores of GrabCut with SGAN is greater than just using HNN with OG images. Considering the simplicity of GrabCut, this indicates that focusing attention on improving data quality can sometimes yield larger gains than simply applying more powerful, but costly, segmentation methods. As a result, despite being somewhat neglected in the field, focusing attention on data enhancements can be an important means to push segmentation performance further. 

Table 2 lists the Dice scores across lesion types when using HNN on OG and SGAN images. Notably, the segmentation performance over all categories is improved with SGAN images, with abdomen and liver exhibiting the largest improvement. These lesion categories may benefit the most due to their low contrast and blurred boundaries compared to the surrounding soft tissue.

\section{Conclusions}
We propose an SGAN method to enhance CT images to improve lesion segmentation performance. SGAN divides the task it into two sub-tasks: the first GAN denoises the original CT image while the second generates a high quality image with higher resolution, enhanced boundaries, and higher contrast. Experimental results on the DeepLesion dataset test segmentation performance when GrabCut and HNN are applied on OG and enhanced images. Results demonstrate that SGAN is more effective than four other enhancement approaches, including using a single GAN, in yielding improved segmentation performance, with HNN + SGAN achieving the best performance. Most notably, Grabcut + SGAN outperformed HNN trained on the OG images, despite the latter having orders of magnitude more parameters. This demonstrates that focusing on dataset processing is a crucial research direction in medical imaging analysis. 
\vspace*{0.5\baselineskip}

\noindent\textbf{Acknowledgments.}
This research was supported by the Intramural Research Program of the National Institutes of Health Clinical Center and by the Ping An Insurance Company through a Cooperative Research and Development Agreement. We thank Nvidia for GPU card donation.
{
	\bibliographystyle{splncs}
	\bibliography{egbib}

\begin{thebibliography}{10}

\bibitem{tang2018semi}
Tang, Y., Harrison, A.P., et~al.:
\newblock Semi-automatic recist labeling on ct scans with cascaded
  convolutional neural networks.
\newblock arXiv:1806.09507 (2018)

\bibitem{jin2018ct}
Jin, D., Xu, Z., et~al.:
\newblock Ct-realistic lung nodule simulation from 3d conditional generative
  adversarial networks for robust lung segmentation.
\newblock arXiv:1806.04051 (2018)

\bibitem{cai2018accurate}
Cai, J., Tang, Y., et~al.:
\newblock Accurate weakly-supervised deep lesion segmentation using large-scale
  clinical annotations: Slice-propagated 3d mask generation from 2d recist.
\newblock arXiv:1807.01172 (2018)

\bibitem{massoptier2008new}
Massoptier, L., Casciaro, S.:
\newblock A new fully automatic and robust algorithm for fast segmentation of
  liver tissue and tumors from ct scans.
\newblock European Radiology \textbf{18}(8) (2008)  1658

\bibitem{christ2016automatic}
Christ, P.F., Elshaer, M.E.A., et~al.:
\newblock Automatic liver and lesion segmentation in ct using cascaded fully
  convolutional neural networks and 3d conditional random fields.
\newblock In: MICCAI. (2016)  415--423

\bibitem{dabov2007image}
Dabov, K., Foi, A., et~al.:
\newblock Image denoising by sparse 3-d transform-domain collaborative
  filtering.
\newblock IEEE TIP \textbf{16}(8) (2007)  2080--2095

\bibitem{zhang2017beyond}
Zhang, K., Zuo, W., et~al.:
\newblock Beyond a gaussian denoiser: Residual learning of deep cnn for image
  denoising.
\newblock IEEE TIP \textbf{26}(7) (2017)  3142--3155

\bibitem{goodfellow2014generative}
Goodfellow, I., Pouget-Abadie, J., et~al.:
\newblock Generative adversarial nets.
\newblock In: NIPS. (2014)  2672--2680

\bibitem{ledig2016photo}
Ledig, C., Theis, L., et~al.:
\newblock Photo-realistic single image super-resolution using a generative
  adversarial network.
\newblock In: CVPR. (2017)  4681--4690

\bibitem{yan2018deep}
Yan, K., Wang, X., et~al.:
\newblock Deep lesion graphs in the wild: relationship learning and
  organization of significant radiology image findings in a diverse large-scale
  lesion database.
\newblock In: CVPR. (2018)  9261--9270

\bibitem{shi2016real}
Shi, W., Caballero, J., et~al.:
\newblock Real-time single image and video super-resolution using an efficient
  sub-pixel convolutional neural network.
\newblock In: CVPR. (2016)  1874--1883

\bibitem{ioffe2015batch}
Ioffe, S., Szegedy, C.:
\newblock Batch normalization: Accelerating deep network training by reducing
  internal covariate shift.
\newblock In: ICML. (2015)  448--456

\bibitem{he2015delving}
He, K., Zhang, X., et~al.:
\newblock Delving deep into rectifiers: Surpassing human-level performance on
  imagenet classification.
\newblock In: ICCV. (2015)  1026--1034

\bibitem{agustsson_2017_cvpr_workshops}
Agustsson, E., Timofte, R.:
\newblock Ntire 2017 challenge on single image super-resolution: Dataset and
  study.
\newblock In: CVPRW. (2017)  1122--1131

\bibitem{simonyan2014very}
Simonyan, K., Zisserman, A.:
\newblock Very deep convolutional networks for large-scale image recognition.
\newblock arXiv:1409.1556 (2014)

\bibitem{deng2009imagenet}
Deng, J., Dong, W., et~al.:
\newblock Imagenet: A large-scale hierarchical image database.
\newblock In: CVPR. (2009)  248--255

\bibitem{kingma2014adam}
Kingma, D.P., Ba, J.:
\newblock Adam: A method for stochastic optimization.
\newblock arXiv:1412.6980 (2014)

\bibitem{rother2004grabcut}
Rother, C., Kolmogorov, V., et~al.:
\newblock Grabcut: Interactive foreground extraction using iterated graph cuts.
\newblock In: ACM TOG. (2004)  309--314

\bibitem{xie2015holistically}
Xie, S., Tu, Z.:
\newblock Holistically-nested edge detection.
\newblock In: ICCV. (2015)  1395--1403

\end{thebibliography}
}

\end{document}